\documentclass{article}
\usepackage{ijcai21}

\usepackage{times}
\usepackage{soul}
\usepackage{url}
\usepackage[hidelinks]{hyperref}
\usepackage[utf8]{inputenc}
\usepackage[small]{caption}
\usepackage{graphicx}
\usepackage{amsmath}
\usepackage{amsthm}
\usepackage{booktabs}
\usepackage{algorithmic}
\usepackage{xcolor}
\usepackage[ruled,noend]{algorithm2e}
\title{Advising Agent for Service-Providing Live-Chat Operators \\
}

\author{
Aviram Aviv$^3$
\and
Yaniv Oshrat$^1$\and
Samuel A. Assefa$^{2}$\and
Tobi Mustapha$^{2}$\and
Daniel Borrajo$^2$\thanks{On leave from Universidad Carlos III de Madrid (consultant).}\and
Manuela Veloso$^2$\thanks{On leave from Carnegie Mellon University.}\and
Sarit Kraus$^1$\
\affiliations
$^1$Department of Computer Science, Bar-Ilan University\\
$^2$JP Morgan AI Research, New York\\
$^3$Department of Mathematics, Bar-Ilan University\\
\emails
aaviv10@gmail.com,
oshblo@zahav.net.il,
samuel.a.assefa@jpmorgan.com,
tmusta78@gmail.com,
daniel.borrajo@jpmorgan.com,
manuela.veloso@jpmorgan.com,
sarit@cs.biu.ac.il\
}

\begin{document}

\maketitle
\begin{abstract}
Call centers, in which human operators attend clients using textual chat, are very common in modern e-commerce. Training enough skilled operators who are able to provide good service is a challenge. We suggest an algorithm and a method to train and implement an assisting agent that provides on-line advice to operators while they attend clients. The agent is domain-independent and can be introduced to new domains without major efforts in design, training and organizing structured knowledge of the professional discipline. We demonstrate the applicability of the system in an experiment that realizes its full life-cycle on a specific domain and analyse its capabilities.   
\end{abstract}

\section{Introduction}

In modern e-commerce, many business services are provided via the Internet. Not only new web-oriented enterprises use this option, but traditional ones as well have moved relevant services to the digital medium. For example, banks are increasingly closing their physical branches and moving services, formerly provided only face-to-face, to the internet~\cite{nam2016internet}. There are many actions that customers can perform by themselves via the Internet, without human intervention, either by a self-service application or using a conversational chatbot. But when customers want to perform actions that do not have an online solution yet, or when they fail to do it by themselves, they still need to approach the bank's customer service and seek human help.

There are several communication channels between the customers and the call center employees (operators). The first method is using a telephone call. This method gives the customer the full attention of someone capable of helping, but at the same time it forces the operator to wait for the customer's reactions. In many of these calls, the customer needs to perform actions he is not familiar with, making the operator wait idly for the customer to finish. Since the operator attends only one customer at a time, this approach wastes time that could be better utilized. With the rise of the Internet, another approach for call centers emerged, using a text-based chat service. This method obviates the constraint of giving one customer the full attention of the operator, as it parallels the service. Instead of talking with one customer at a time, the operator interacts with 2-4 customers in parallel, switching between customers instead of waiting for the customer's reactions. 
 
While this approach has its advantages, it also raises some challenges for the human operator to deal with; as the number of tasks that the operator has to perform simultaneously grows, so may her stress. The operator also needs to prioritize the tasks, keep track of each individual's information while assisting different clients, and provide help without making any client wait too long.


We propose to mitigate these challenges by assisting the human operator by creating an \textbf{advising agent}. This kind of agent works alongside the operator during the chat session and suggests on-line advice to help the operator deal with the situation.
To the best of our knowledge, we are the first to use
an advising agent to cope with these challenges. However, building an advising agent and training it to the specific service domain can be a long and expensive process that requires both domain expertise and system engineering knowledge.




\section{Related Work}

 

In recent years, many companies have started to develop chatbots for the task of customer service~\cite{nuruzzaman2018survey}. There is much work regarding the design and deployment of such bots in various domains~\cite{okuda2018ai}. It is evident that at the current phase chatbots cannot fully replace human workers, and when a bot detects it cannot help the customer, it refers the customer to human help. Li {\it et al.}~\shortcite{li2020conversation} and Liu {\it et al.}~\shortcite{liu2020time} explore the challenge of  detecting this kind of situation in various domains.

 
\subsection{Agents that Support Human Actions}

 Intelligent agents that support humans in their complex activities need to be able to predict the user's behavior and decisions~\cite
 {rosenfeld2016online,kraus2016human,rosenfeld2018predicting}. This is a difficult task because of an extensive set of factors that influence human decision-making and behavior,
 including inherent differences between individuals and groups of individuals~\cite{bruine2007individual}, which make the prediction of an individual's decisions and behavior even more challenging.
 
There are numerous previous predictions of human behavior and decisions in agent-human interactions. Azaria {\it et al.}~\shortcite{Azaria2015} developed CARE (automobile Climate control Advisor for Reducing Energy consumption) -- an agent that uses two models,
finds a compromise between them and offers it to the driver, who chooses whether or not to accept it.

Rosenfeld {\it et al.}~\shortcite{rosenfeld2017intelligent} developed automated agents that can assist a single operator to better manage a group of robots,
showing that an agent can significantly improve the performance of a team of an operator and ten low-cost mobile robots. 
This work also compared two approaches of advice: one looks for the advice that will have the best impact on the current situation, and the other searches for advice that will lead to better results in the near future. 
 
\subsection{Method of Advice Provision}
 
When it comes to advising in repeated human interaction environments, several methods were used in the literature.
Rosenfeld {\it et al.}~\shortcite{rosenfeld2017intelligent} directly estimated the reward for every possible advice and recommended a piece of advice that maximizes the reward the user will get if the provided advice is chosen. Elmalech {\it et al.}~\shortcite{elmalech2015} suggested that the agent will try to find a compromise between maximizing rewards and user acceptance.
The common ground of these advising agents is that they all advise in order to maximize a certain reward function. This kind of advising mechanism tends to recommend non-intuitive advice that the operators tend to reject, making them ineffective~\cite{carroll1988}.

\subsection{Learning How to Provide Advice}

In recent years, companies began keeping records of their workers' actions and their interactions with the customers due to low digital storage costs~\cite{katal2013}. This accumulation of information was mostly used for basic performance analysis but with the improvement of machine learning capabilities this abundance called for new uses\cite{Esposito2015}.

Argall {\it et al.}~\shortcite{Argall2009} coined the term “learning from demonstration" (LfD) for the practice of using human actions and decisions as the base of the learning process (though there are several other terms in the literature).
The LfD approach is used in a large variety of fields (e.g, \cite{Wong2018,Floyd2017}).
In our context, Levy and Sarne~\shortcite{levy2016} combined LfD and advice provision as they created an agent that used the way people act in a specific scenario in order to guess what they would do in similar situations.


Even though there are many ways in the literature to generate conversational data~\cite{crook2017,majumdar2019generating,madotto2020learning}, we focus on using human-human conversations for the learning process of the agent because they better reflect the real-life scenario~\cite{Williams2007} and hopefully help in generating more intuitive advice.
\section{Modus Operandi and Life-Cycle of the Agent}
\label{sec:modus_operandi}

Our research goal is to develop a domain-independent, data-driven algorithm for an automated system that will assist in the operators' training process and daily activities, will help new and experienced operators, and will advise the operators about possible actions and relevant information during textual chat interactions with customers in real-time. Implementing the automated system (i.e., the agent) in a call center will reduce daily workload and improve the interaction with the customers from the human operator's point of view. It will improve the system's service efficiency and reduce the time needed to help each customer. 

\subsection{Agent's Life-Cycle}
The process of building an advising-agent for a new domain is performed in three phases, as follows: 
\begin{enumerate}
    \item \textbf{The Apprentice Phase (Phase 1)} -- experienced human operators serve human customers regarding the new domain of service.
    The operators tag the information they find important (in real time or afterwards).
    The collected data is fed to the learning process (as detailed in Section~\ref{sec:learning_process}). There is no agent assistance in this phase. Section~\ref{sec:apprectice_phase} elaborates about the experimentation of this phase.
    \item \textbf{The Novice-Advisor Phase (Phase 2)} -- this phase contains both data collection (for the improvement of the agent's capabilities) and service to clients: the agent works alongside a non-experienced human operator who attends clients, and it simultaneously advises and learns. For advising the human operator, the agent uses the tagging to predict the message that the operator should send or an action it should perform, and offers it to the operator. The operator may use this advice or not, as suits her.
    
    In addition, the data collected in this phase may be fed into the agent's machine learning model in order to improve its tagging and advising capabilities. This feeding may be performed daily, weekly, monthly or in any batch form that is suitable to the managers of the service. Additional details regarding this phase are presented in Section~\ref{sec:novice_advisor_phase}.
    
    \item \textbf{The Expert-Advisor Phase (Phase 3)} -- The agent works alongside a non-experienced human operator and provides advice based on former tags and a learned model. The agent is not engaged in further learning, since its capabilities have already reached an adequate level. This phase is the final and steady state of the agent in the current domain. 
\end{enumerate}

A rollback from Phase 3 to previous phases may be performed if new circumstances emerge in reality. The system can be returned to Phase 1 or to Phase 2 (according to managers' preferences), collect data (i.e. tagged chat conversations) and feed them to the machine learning model. When reaching the desired level of advice, the agent may be advanced again to Phase 3, and so forth. 

\subsection{The Learning Process}
\label{sec:learning_process}
In order to provide advice, the agent relies on a predictive model learned from observations of the domain: Operators conduct chat sessions with clients and attend to their needs. During the chat sessions, the operators tag the vital information they used to reach the satisfactory outcomes (there is no tagging in advance). Each session's tag list is turned into an \textbf{information vector}. Each time a new tag is added, the vector's current version is saved to be used later in the learning process as an information vector.



\subsubsection{Building the Information Vector}
We build the information vector as follows:
At first, we take the $n$ most common features that operators marked in the data and sort them in alphabetical order (the label list).


Notation remark: We write $X_{t}$ as the information vector after $t$ pieces of information (that is, $X$ at time $t$), and $X[t]$ for the value of $X$ at index $t$. Whenever we mention tag $i$, we refer to the value of the tag list at index $i$. 

\begin{figure}[ht]
\centering
\includegraphics[width=0.48\textwidth]{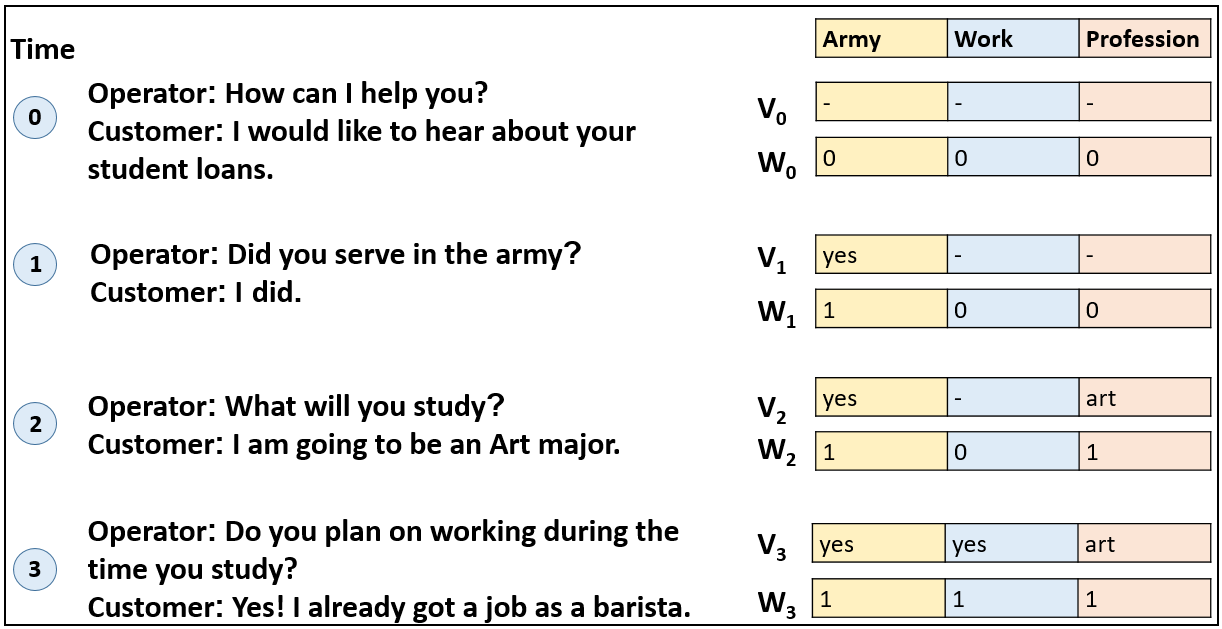}
\caption{\label{fig:chat_example} The process of building an information vector as the chat between an operator and a client proceeds.}
\end{figure}

We define two vectors of size $n$.
The first one is:

{\scriptsize
$$
    V[i]:=
\begin{cases}
    word,& \text{{\footnotesize if a known word was tagged as label i}} \\
    "unknown",& \text{{\footnotesize if an unknown word was tagged as label i}} \\
    "-",              & \text{{\footnotesize if no input was labeled as label i}}
\end{cases}
$$
}

and the second one is:

$$
    W[i]:=
\begin{cases}
    1,& \text{if there is an input labeled as label i} \\
    0,              & \text{otherwise}
\end{cases}
$$
the vector $X$ is a concatenation of ($V,W$). Figure \ref{fig:chat_example} demonstrates the building of the information vector.

\subsubsection{Advice Types}
There are three types of advice that we wish to provide: (1) Topic acquisition -- questions the operator should ask the client in order to acquire information she needs in order to help him; (2) Resolution -- data the operator should provide to the client as a response to his query; and (3) Useful information -- data the operator may need in order to provide answers, such as relevant websites, calculations, etc. 

\subsubsection{Advice Providing Process}
During the call with the customer, whenever the operator uses a website operation or finds out new information about the customer, useful information is marked under a suitable label (the feature), and the customer information vector $X$ is updated accordingly.

For each advice type $i$ of the three mentioned above, advice is provided by taking the feature vector $X_{t}$ and inserting it into a machine learning module $F_{i}$ that tries to find the best set of advice $A$ for the current situation ($A_{t}$). The module uses $k$ pairs ${D_1 \dots D_{k}}$ of past experiences $D_j$=($X_j$,$A_j$), where $X_j$ is an information vector at time $j$ and $A_j$ is the respective set of advice, in order to find a set that maximizes the chance to be the most used set of advice in the past similar situations:
 $P(A_{t}=A|X_t,D_1,...,D_k)$

For the learning algorithm, we wanted to find an algorithm with the ability to work efficiently on several domains and handle messy and conflicting data. The first model that came to mind was Random Forest, a model that works well but cannot fully utilize the vast amount of data usually available in such domains. To deal with this problem, we thought of using neural networks. That idea also worked pretty well, but an architecture that works on one domain might fail to learn on another. 
With all that in mind, we decided to combine them as an ensemble method of neural networks~\cite{hansen1990neural} where each network takes the information known about a customer at a certain time and outputs the recommended set of advice for the situation. 
Each network in the ensemble was trained on a subset of the data and had a random number of layers of an arbitrary length, as shown in Algorithm 1.

\begin{algorithm}[h]
\SetAlgoNoLine
\caption{Training the ensemble:}
\KwResult{a list of trained neural networks}
nets=$\emptyset$\\
\While{length(nets) \textless  ensembleSize}{
    num=GetRandomNumber()\\
    \eIf{num \textgreater 0.5}{
       trainSet=getRandomSubSet(trainData) 
        }{
        trainSet=getBalancedSubSet(trainData) 
        }
    net=GenerateRandomNeuralNetwork()\\
    train(net,trainSet)\\
    $P_{net}$=accuracy(net,testData)\newline 
    \If{$ P_{net}$ \textgreater $P_{threshold} $}{
        nets $ \xleftarrow[]{\text{add}}$ net
    }
}
\end{algorithm}

The final set of advice is chosen using a majority voting variation (as shown in Algorithm 2). We also tested this method against other variations of Random Forest (LGBM ~\cite{ke2017lightgbm} and regular Random Forest) and other crowd related algorithms (SVM and KNN). This method outperformed the others in an 80:20 cross-validation where the target label needed to be in the top 2 recommendations (the ensemble reached 87\% accuracy, regular and gradient boosted Random Forests with 84\%, KNN with 83\%, neural network with 77\% and SVM with 70\%). We chose this metric because there can be a large variation based on the operator's preferences, even in a small amount of data.

\begin{algorithm}[h]
\SetAlgoNoLine
\KwResult{Final recommendations}

results=$\emptyset$\\
X=getData()\\
\For{net in ensemble}{  
results$\xleftarrow[]{\text{add}}$ prediction(X)\

} 
bestOptions=twoMostCommonOptions(results)\\
finalRecommendations=$\emptyset$\\
\If{rankOf(bestOptions[0]) \textgreater firstOptionThreshold }{
finalRecommendations $\xleftarrow[]{\text{add}}$ bestOptions[0]\

}
\If{rankOf(bestOptions[1]) \textgreater secondaryOptionThreshold }{
finalRecommendations $\xleftarrow[]{\text{add}}$ bestOptions[1]\

}
return finalRecommendations\\
\caption{Using the ensemble:}
\end{algorithm}

As we can see, the agent can recommend one set, recommend  a combination of two sets, or remain silent (when $\emptyset$ is chosen or when finalRecommendations is empty).

\section{Experiment}
\label{sec:experiment}
Our experiment was designed to test whether implementing the suggested system will yield better performance of the operators. For this purpose we chose a domain, set up a working environment and recruited participants to play the roles of operators and clients in various configurations. At the end of each session the operators filled out questionnaires to quantify their opinions regarding different aspects of the performance. We hereby describe the setup and course of the experiment. The results will be presented in Section~\ref{sec:Results}.

The domain on which we chose to perform our experiment is students loans in the US. Customers who are interested in understanding their options in getting such loans, either for themselves or for their relatives (usually a son or a daughter), call the information center and chat with the operator. In some cases, the customers know what the relevant data is, and they can provide it to the operator right away. Nevertheless, in many cases the customers are not familiar with the parameters that define their entitlement to a loan, and they should be guided. For example, in the US men are required to register in the Selective Service System in order to be entitled to a federal loan. Many loaners are not aware of this requirement, and informing them of it, or of other parameters that might affect their ability to get a loan of the sum they need, is very beneficial. Good service by the operator should clarify these issues in order to enable the customer to exhaust his rights. Hence, there is much room for accurate advice to the operator in this process.
Since Phase 3 in our model is the steady state working mode, we performed our field experience on phases 1 and 2 which implement the building of the model. 

\subsection{Phase 1 -- the Apprentice Phase}
\label{sec:apprectice_phase}
As mentioned above, the goal of this phase is to provide the agent with labeled data regarding our domain by listening to sessions in which experienced human operators chat with clients. This phase was implemented in our experiment by recruits that played the operators and the customers. The operators were thoroughly briefed and trained about the domain and the service they should provide to customers. At the end of this preparation stage the recruits were at the level of an experienced operator in the domain of student loans. The customer received storyboards, each with character information (profession, university, savings, financial status etc.) and objectives to achieve (loan options, pre-specified information about the loans etc.).

\begin{figure}[t]
\centering
\includegraphics[width=0.48\textwidth]{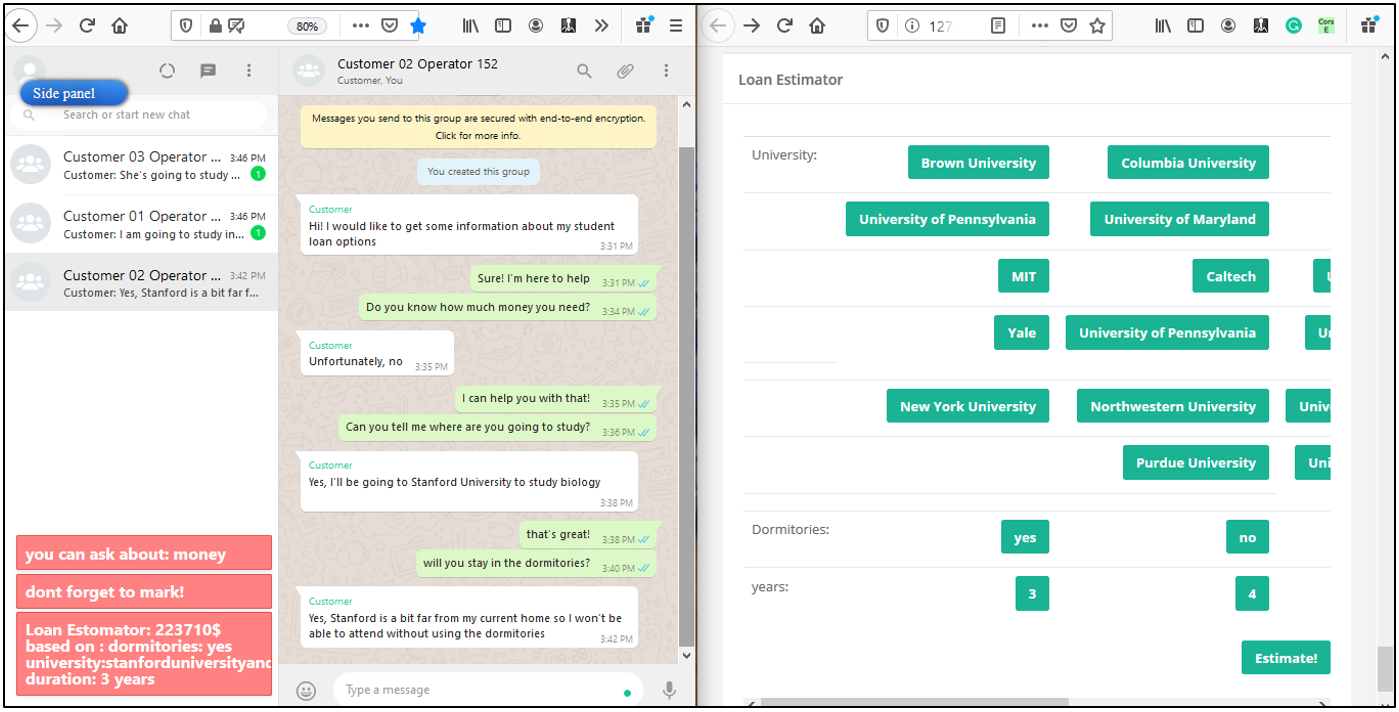}
\caption{\label{fig:screenshot} A screenshot of the operator interface. The left-hand side is the "WhatsApp" chat interface that enables the operator to switch easily between clients. On the right-hand side the operator can tag information in the conversation and use calculators. The red rectangle (bottom-left) presents the advice that the agent offers to the operator.}
\end{figure}

The chat between the customers and the operators was performed using a textual chat application. The operators used a computer where half of the screen is a “WhatsApp web" interface with a special browser extension that knows when the operator switches between two conversations (as a single operator attended 2-3 customers simultaneously), and allows the operator to mark words. The other half of the screen shows a website which presents relevant information and enables the operator to perform common calculations by clicking on pre-defined buttons (see Figure~\ref{fig:screenshot}). 

The subjects played multiple client-operator sessions. In these sessions there was no participation of an assisting agent, and only the human operators and the human clients took part. During the sessions, in addition to collecting relevant information from the clients and answering clients' questions, the operators also tagged phrases in the chat. They were asked to tag (by marking words on the screen) any information that they considered relevant to the loan issue. Each tag contained a category (e.g. university name) and a value for that category (e.g. UCLA, MIT, Columbia University). Operators were neither told nor limited regarding what categories of tags they can mark. They saw what categories were tagged and used before, but did not see their values. They could add additional tags as needed.

In this phase of the experiment, 4 subjects took part as operators and one subject played the clients. In total there were 76 sessions, and in each of them a single operator attended 2-3 simulated clients.

    
\subsection{Phase 2 -- the Novice-Advisor Phase}
\label{sec:novice_advisor_phase}
The goal of the Novice-Advisor Phase is twofold: To assist operators in their work, as well as to collect additional data for the improvement of the agent. In the experiment, our main goal was to evaluate the helpfulness of the agent we built in Phase 1.

In our experiment this phase was implemented using recruits from the AI course for undergraduate students in XXX University (full name not mentioned due to anonymity) as clients, and paid recruits from the general population as operators or clients. Each operator played two sessions: one with an agent's assistance and the other without it. Half of the operators played the assisted session first and the unassisted session second, and the other half vice versa. 

At the end of each session, we asked the participants who played as operators to 
fill out a NASA-TLX questionnaire~\cite{hart1988development}, which is an assessment tool for comparing the workload of different tasks (a summary of the NASA-TLX questionnaire can be found in the supplementary material). These opinions were analyzed in order to evaluate the performance of the agent and its contribution to the performance of the operators. The findings of the analysis are presented in Section~\ref{sec:Results}. 

At this point we had 23 operators who played 46 sessions: 15 of the 23 operators attended 2 clients simultaneously and the remaining 8 operators attended 3 clients simultaneously. 
The tagging of the text was done manually during the sessions by the operators.

\subsubsection{The Tagging Problem}
The tagging of the chat conversations is essential for the agent in order to follow the line of conversation and provide proper advice. Our preliminary design was to tag the chat by the human operator, in real-time, during the session. Unfortunately, we found out that the operators of Phase 2 managed to perform the tagging well while attending 2 clients, but when they needed to attend 3 clients simultaneously the workload was too heavy, and they could not tag the conversation properly; as the session proceeded, there was much less tagging or none at all. As a result the ability of the agent to provide advice weakened. This situation called for a change.

In order to perform good tagging even in loaded sessions, we introduced an automated tagging mechanism. We took the raw data in real-time and made the agent use it directly, a common notion in goal-oriented dialogue systems, and chatbots in general~\cite{serban2015}. We used a machine learning module that follows the messages in real-time and outputs annotations for the advising agent. The module that we chose is a combination of two sub-models, as follows: We denote a network consisting of a BERT~\cite{devlin2019bert} embedding layer with a linear layer on top as a \textbf{BERTLL}. At first, a BERTLL predicts what information categories the message may contain. For each category that the first model predicted, another BERTLL predicted what tag the message contains. We chose to use this combination after it reached a maximum $F_1$ score of $0.72$ and was seen to generalize well in practice. It also outperformed a gradient boosted Random Forest (that reached an $F_1$ score of $0.7$), a single BERTLL for all the labels (that reached a maximum $F_1$ score of $0.5$)
and a large variety of neural network based models that were far from reaching a $0.5$ $F_1$ score. Implementing the automated tagging mechanism relieves the operator from the tagging task, and enables her to concentrate on the sole task of attending the clients.

Another improvement in the experiment method (relative to the original design) was the introducing of bots as clients in this phase: Instead of human subjects playing the role of clients, we deployed bots that were built using a combination of two strategies: a rule-based approach, and a learning approach. The first approach followed the spirit of early chatbots, such as Eliza~\cite{eliza}. Based on the previous interaction with the operator, the bot would randomly generate answers to operator's questions, or questions for asking the operator. The second approach used BERT to learn how to perform the interaction. We found out that these two models (knowledge-based and learning-based) complemented each other quite well in overcoming their respective disadvantages. This change was made because the use of bots instead of human subjects made the experiment much easier to perform, since we needed to recruit and to brief only the operators, and the influence of the agent on the clients was not examined in this research. 

After implementing the aforementioned changes, we performed the experiment of Phase 3, this time with each operator attending 3 clients simultaneously. In this improved design we did not encounter an excessive load on the operators since the tagging was done automatically by the agent.
We had 14 operators playing two sessions each (again half of the operators played the assisted session first and the unassisted session second, and the other half vice versa). Together with the 15 operators who attended 2 clients simultaneously, we had 29 operators, and each of them played 2 sessions. The demographic data regarding the subjects in the experiment is presented in the supplementary material.

\section{Results}
\label{sec:Results}
\subsection{Operators' Opinions} 
\label{sec:Results_operators_opinions}
The participants who played the role of operators filled out TLX questionnaires. There are six categories in this questionnaire which are factored into the total TLX grade (as elaborated on in the complementary material). We compared the grades regarding sessions that were played with the agent's assistance to the grades regarding sessions that were played without the agent's assistance. We specified five categories:

\begin{enumerate}
    \item First Session - only the first session of every operator.
    \item Total Sessions - all of the sessions (both the first and the second) of all of the operators.
    \item First with agent - only the sessions of operators whose first session was played with an agent's assistance.
    \item First without agent - only sessions of operators whose first session was played without an agent's assistance.
\end{enumerate}

We also divided the data into sessions with two simultaneous customers and sessions with three simultaneous customers, and looked at these groups separately.

\begin{figure}[t]
\centering
\includegraphics[width=0.48\textwidth]{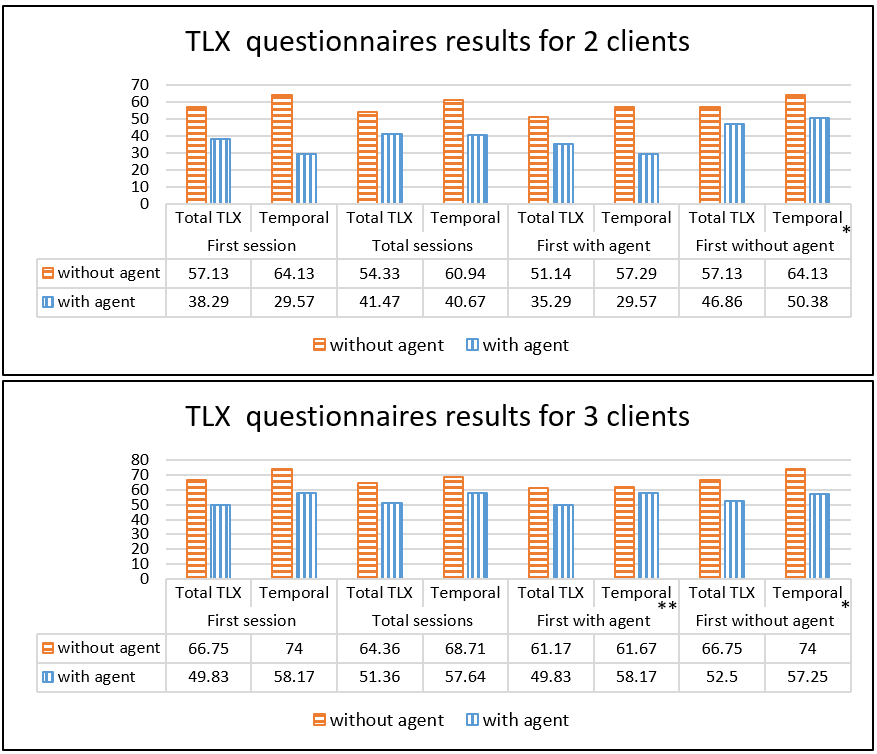}
\caption{\label{fig:tlx_data} TLX questionnaires data (lower is better).}
\end{figure}

Figure~\ref{fig:tlx_data} presents the results of the TLX questionnaires in the aforementioned categories for sessions in which two clients were attended simultaneously by a single operator. It can be seen that total workload (Total TLX) and temporal demand \textbf{decreased in all cases} when having the agent by the operator's side as compared to not having the agent. All data presented in Figure~\ref{fig:tlx_data} are statistically significant ($p<0.05$), except columns marked with an asterisk which have $p<0.07$, and columns with two asterisks ($p<0.5$).

\subsection{Time Performance}
\label{sec:Results_time_performance}
\begin{figure}[b]
\centering
\includegraphics[width=0.48\textwidth]{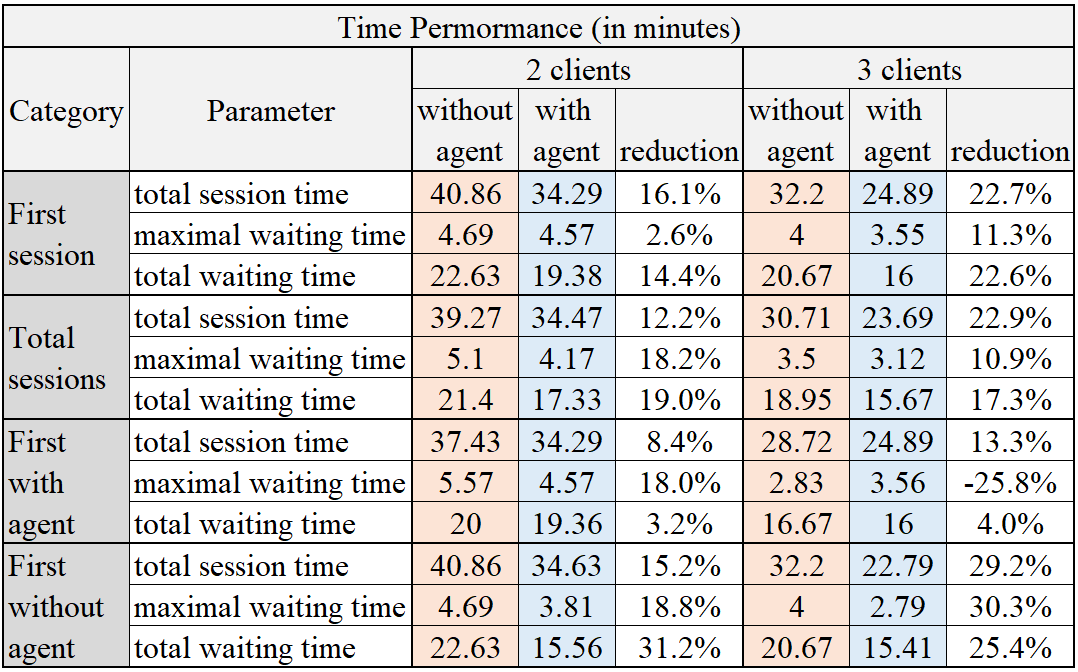}
\caption{\label{fig:time_performance} Time performance data (in minutes, decimal notation).}
\end{figure}

We presumed that a good performance of the agent would be manifested in providing the service in shorter time, and with less idle time during the session.
Figure~\ref{fig:time_performance} presents the time performance data of the operators in three categories:
\begin{enumerate}
    \item Total session time - the average length of a full session, including clients' time, operator's time and idle time.
    \item Maximal waiting time - the maximal time a client had to wait for an operator's response.
    \item Total waiting time - the average total time a client spent waiting for an operator's responses during a session.
\end{enumerate}

In all categories (except one) the times are shorter when the agent assisted the operator than when it did not, and in most of the categories the reduction is greater than 10\%. It implies that the use of an agent alongside the operator reduces the time needed for the session in general and the time spent by the client idly waiting for the operator to respond.   Nevertheless, the data was not found statistically significant in most of the categories.


\subsection{Learning Effectiveness of Phase 2}
The Apprentice Phase (Phase 1) is, naturally, crucial to the building of the preliminary knowledge base of the agent. Nevertheless, we wondered whether the Phase 2 actually improves the capabilities of the agent, or if it is superfluous and we may skip it and go straight to the final stage (Phase 3). In order to answer this question we compared the performance of the tagging model in two configurations: The first one was based on data collected in Phase 1 only, while the second one was based on data collected in both Phase 1 and Phase 2. We found that the performance of the second model (precision -- $78\%$, recall -- $75\%$, $F_1$-score -- $72\%$) was better than the performance of the first model (precision -- $65\%$, recall -- $60\%$, $F_1$-score -- $58\%$). This result indicates that although the data of Phase 1 alone suffices to provide basic assistance to human operators, expanding it with the data of Phase 2 significantly improves the tagging capability and, as a result, the quality of the agent's performance.      

\section{Conclusions, Discussion and Future Work} 
In this paper we introduced an algorithm and a method to implement an advising agent that assists operators who attend clients in a call center using chat conversations. The main advantage of this method is its adaptability -- the agent is domain-independent and can be fitted to a new domain with relatively little effort and short time. Training the agent does not require long design, domain analysis or rule-definition. In an existing human call center, the agent only needs tagged conversation logs with clients and experienced human operators in order to build all of the required knowledge.

Integrating the results of the role-playing experiment, we see that operators who were assisted by the agent enjoyed a lower cognitive load in attending their clients, with less effort and less time-pressure. Time is used in a more efficient way, as sessions are shorter and less time is spent on idle waiting. This trend is evident both in the objective measure of time to perform a mission (Section \ref{sec:Results_time_performance}) and in the subjective views of the participants who played as operators (Section \ref{sec:Results_operators_opinions}).  

There are several issues that still need to be examined. One such issue is an optimization of the process of adjusting the agent to a new domain. We found that the Novice-Advisor Phase (Phase 2) indeed improves the performance of the agent, and therefore the 3-stage process that was suggested is justified. However, the optimal conditions for switching from Phase 2 to Phase 3 still need to be determined.

Another issue is the possibility for an operator to attend to a larger number of clients simultaneously when having the agent's assistance. We performed experiments attending 2 and 3 clients because this was the customary situation for a single operator. However, an operator might be able to attend more than 3 clients simultaneously by having an agent by her side. The feasibility of this option should be tested as well.

It should be noted that this research study was designed to examine the effects of the assisting agent on the assisted operators. A differently designed experiment may explore the influence of the agent on the service from the clients' perspective.

\section*{Acknowledgement}
This paper was prepared for informational purposes in part by
the Artificial Intelligence Research group of JPMorgan Chase \& Co\. and its affiliates (``JP Morgan''),
and is not a product of the Research Department of JP Morgan.
JP Morgan makes no representation and warranty whatsoever and disclaims all liability,
for the completeness, accuracy or reliability of the information contained herein.
This document is not intended as investment research or investment advice, or a recommendation,
offer or solicitation for the purchase or sale of any security, financial instrument, financial product or service,
or to be used in any way for evaluating the merits of participating in any transaction,
and shall not constitute a solicitation under any jurisdiction or to any person,
if such solicitation under such jurisdiction or to such person would be unlawful.

\appendix
\section{The NASA TLX Questionnaire}

The NASA TLX questionnaire~\cite{hart1988development} is an assessment tool for comparing the workload of different tasks. While it is subjective to the worker's perspective, it is a commonly used benchmark in many papers~\cite{hart2006nasa,seker2014using,miyake2020mental}. The final score is calculated using grades the worker gave to the following categories:

Mental demand - How mentally demanding was the task?

Physical demand - How physically demanding was the task?

Temporal Demand - How hurried or rushed was the pace of the task?

Performance - How successful was the worker in accomplishing what he/she was asked to do?

Effort - How hard did the worker have to work to accomplish that level of performance?

Frustration - How insecure, discouraged, irritated, stressed, and annoyed was the worker?

All of the grades above are ranged from 0 to 100, where 0 means the parameter didn't affect the workload, and 100 means that the parameter heavily contributed to the workload. In the context of our experiment, as the goal is to ease the load the operators feel during their work, "lower is better" -- lower grades mean that the load felt by the operators is lower.

\section{Subjects Demographic Data}
\begin{figure}[t!]
\centering
\includegraphics[width=0.48\textwidth]{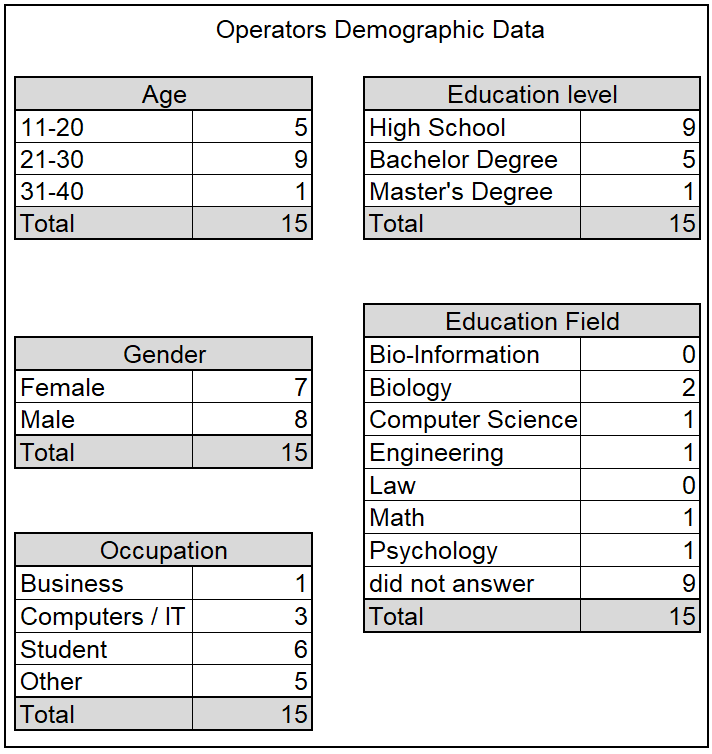}
\caption{\label{fig:operators_data_2} Demographic data of the subjects who played the role of operators in the first stage of the experiment -- each operator attended 2 clients simultaneously.}
\end{figure}

\begin{figure}[]
\centering
\includegraphics[width=0.48\textwidth]{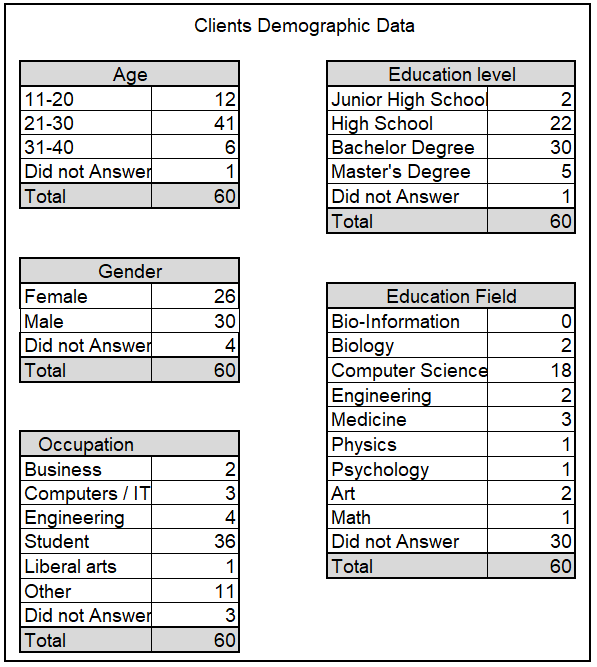}
\caption{\label{fig:clients_data} Demographic data of the subjects who played the role of clients in the the first stage of the experiment. \newline}
\end{figure}

\begin{figure}[t!]
\centering
\includegraphics[width=0.48\textwidth]{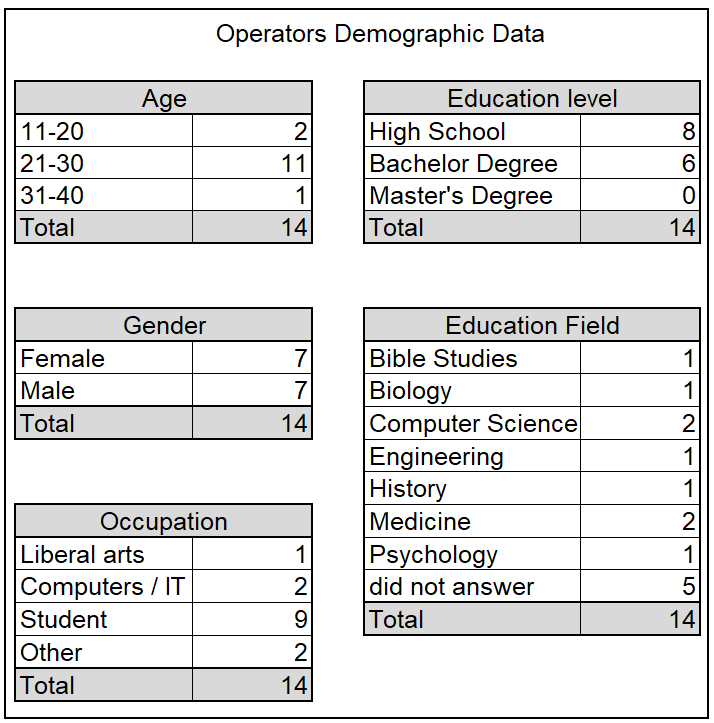}
\caption{\label{fig:operators_data_3} Demographic data of the subjects who played the role of operators in the second stage of the experiment -- each operator attended 3 clients simultaneously .}
\end{figure}

Herewith is the demographic data of the human subjects who took part in Phase 2 of our experiment. As explained in the paper, the experiment was performed in two stages: In the first stage, operators attended 2 clients simultaneously (the operators' demographic data is presented in Fig.~\ref{fig:operators_data_2}), and human subjects played the role of clients (their demographic data is presented in Fig~\ref{fig:clients_data}). In the second stage, operators attended 3 clients simultaneously (these operators demographic data is presented in Fig~\ref{fig:operators_data_3}), and we used bots as clients (therefore there is no relevant demographic data).

\newpage
\clearpage

\bibliographystyle{named}
\typeout{}
\bibliography{shortreferences}

\begin{thebibliography}{}

\bibitem[\protect\citeauthoryear{Argall \bgroup \em et al.\egroup
  }{2009}]{Argall2009}
B.~D. Argall, S.~Chernova, M.~Veloso, and B.~Browning.
\newblock A survey of robot learning from demonstration.
\newblock {\em ROBOT. AUTON. SYST.}, 57(5):469--483, 2009.

\bibitem[\protect\citeauthoryear{Azaria \bgroup \em et al.\egroup
  }{2015}]{Azaria2015}
Amos Azaria, Ariel Rosenfeld, Sarit Kraus, Claudia~V Goldman, and Omer
  Tsimhoni.
\newblock Advice provision for energy saving in automobile climate-control
  system.
\newblock {\em AI Magazine}, 36(3):61--72, 2015.

\bibitem[\protect\citeauthoryear{Bruine~de Bruin \bgroup \em et al.\egroup
  }{2007}]{bruine2007individual}
W.~Bruine~de Bruin, A.~M. Parker, and B.~Fischhoff.
\newblock Individual differences in adult decision-making competence.
\newblock {\em J. PERS. SOC. PSYCHOL.}, 92(5):938, 2007.

\bibitem[\protect\citeauthoryear{Carroll \bgroup \em et al.\egroup
  }{1988}]{carroll1988}
J.~S. Carroll, M.~H. Bazerman, and R.~Maury.
\newblock Negotiator cognitions: A descriptive approach to negotiators'
  understanding of their opponents.
\newblock {\em Organ. Behav. Hum. Decis. Process.}, 41(3):352--370, 1988.

\bibitem[\protect\citeauthoryear{Crook and Marin}{2017}]{crook2017}
Paul~A Crook and Alex Marin.
\newblock Sequence to sequence modeling for user simulation in dialog systems.
\newblock In {\em INTERSPEECH}, pages 1706--1710, 2017.

\bibitem[\protect\citeauthoryear{Devlin \bgroup \em et al.\egroup
  }{2019}]{devlin2019bert}
J.~Devlin, M.Chang, K.~Lee, and K.~Toutanova.
\newblock Bert: Pre-training of deep bidirectional transformers for language
  understanding, 2019.

\bibitem[\protect\citeauthoryear{Elmalech \bgroup \em et al.\egroup
  }{2015}]{elmalech2015}
A.~Elmalech, D.~Sarne, and B.~J. Grosz.
\newblock Problem restructuring for better decision making in recurring
  decision situations.
\newblock {\em AUTON. AGENTS MULTI-AGENT SYST.}, 29(1):1--39, 2015.

\bibitem[\protect\citeauthoryear{Esposito \bgroup \em et al.\egroup
  }{2015}]{Esposito2015}
C.~Esposito, M.~Ficco, F.~Palmieri, and A.~Castiglione.
\newblock A knowledge-based platform for big data analytics based on
  publish/subscribe services and stream processing.
\newblock {\em KNOWL-BASED. SYST.}, 79:3--17, 2015.

\bibitem[\protect\citeauthoryear{Floyd \bgroup \em et al.\egroup
  }{2017}]{Floyd2017}
Michael~W Floyd, JT~Turner, and David~W Aha.
\newblock Using deep learning to automate feature modeling in learning by
  observation: a preliminary study.
\newblock In {\em 2017 AAAI Spring Symposium Series}, 2017.

\bibitem[\protect\citeauthoryear{Hansen and Salamon}{1990}]{hansen1990neural}
Lars~Kai Hansen and Peter Salamon.
\newblock Neural network ensembles.
\newblock {\em IEEE Trans. Pattern Anal. Mach. Intell.}, 12(10):993--1001,
  1990.

\bibitem[\protect\citeauthoryear{Hart \bgroup \em et al.\egroup
  }{1988}]{hart1988development}
Sandra~G Hart, Lowell~E Staveland, et~al.
\newblock Development of {NASA-TLX} (task load index): Results of empirical and
  theoretical research.
\newblock 52:139--183, 1988.

\bibitem[\protect\citeauthoryear{Hart}{2006}]{hart2006nasa}
Sandra~G Hart.
\newblock Nasa-task load index (nasa-tlx); 20 years later.
\newblock In {\em Proceedings of the human factors and ergonomics society
  annual meeting}, volume~50, pages 904--908. Sage Publications Sage CA: Los
  Angeles, CA, 2006.

\bibitem[\protect\citeauthoryear{Katal \bgroup \em et al.\egroup
  }{2013}]{katal2013}
A.~Katal, M.~Wazid, and R.~H. Goudar.
\newblock Big data: issues, challenges, tools and good practices.
\newblock In {\em 2013 Sixth international conference on contemporary computing
  (IC3)}, pages 404--409. IEEE, 2013.

\bibitem[\protect\citeauthoryear{Ke \bgroup \em et al.\egroup
  }{2017}]{ke2017lightgbm}
G.~Ke, Q.~Meng, T.~Finley, T.~Wang, W.~Chen, W.~Ma, Q.~Ye, and T.~Liu.
\newblock Lightgbm: A highly efficient gradient boosting decision tree.
\newblock In {\em ADV. NEUR. IN.}, pages 3146--3154, 2017.

\bibitem[\protect\citeauthoryear{Kraus}{2016}]{kraus2016human}
Sarit Kraus.
\newblock Human-agent decision-making: Combining theory and practice.
\newblock {\em arXiv:1606.07514}, 2016.

\bibitem[\protect\citeauthoryear{Levy and Sarne}{2016}]{levy2016}
Priel Levy and David Sarne.
\newblock Intelligent advice provisioning for repeated interaction.
\newblock In {\em Thirtieth AAAI Conference on Artificial Intelligence}, 2016.

\bibitem[\protect\citeauthoryear{Li \bgroup \em et al.\egroup
  }{2020}]{li2020conversation}
C.~Li, S.~Yeh, T.~Chang, M.~Tsai, K.~Chen, and Y.~Chang.
\newblock A conversation analysis of non-progress and coping strategies with a
  banking task-oriented chatbot.
\newblock In {\em PROC CHI 2020}, pages 1--12, 2020.

\bibitem[\protect\citeauthoryear{Liu \bgroup \em et al.\egroup
  }{2020}]{liu2020time}
Z.~Liu, J.and~Gao, Y.~Kang, Z.~Jiang, G.~He, C.~Sun, X.~Liu, and W.~Lu.
\newblock Time to transfer: Predicting and evaluating machine-human chatting
  handoff.
\newblock {\em arXiv preprint arXiv:2012.07610}, 2020.

\bibitem[\protect\citeauthoryear{Madotto \bgroup \em et al.\egroup
  }{2020}]{madotto2020learning}
A.~Madotto, S.~Cahyawijaya, G.~I. Winata, Y.~Xu, Z.~Liu, Z.~Lin, and P.~Fung.
\newblock Learning knowledge bases with parameters for task-oriented dialogue
  systems.
\newblock {\em arXiv preprint arXiv:2009.13656}, 2020.

\bibitem[\protect\citeauthoryear{Majumdar \bgroup \em et al.\egroup
  }{2019}]{majumdar2019generating}
Sourabh Majumdar, Serra~Sinem Tekiroglu, and Marco Guerini.
\newblock Generating challenge datasets for task-oriented conversational agents
  through self-play.
\newblock {\em arXiv preprint arXiv:1910.07357}, 2019.

\bibitem[\protect\citeauthoryear{Miyake}{2020}]{miyake2020mental}
Shinji Miyake.
\newblock Mental workload assessment of health care staff by nasa-tlx.
\newblock {\em Journal of UOEH}, 42(1):63--75, 2020.

\bibitem[\protect\citeauthoryear{Nam \bgroup \em et al.\egroup
  }{2016}]{nam2016internet}
Kiheung Nam, Zoonky Lee, and Bong~Gyou Lee.
\newblock How internet has reshaped the user experience of banking service?
\newblock {\em KSII Transactions on Internet \& Information Systems}, 10(2),
  2016.

\bibitem[\protect\citeauthoryear{Nuruzzaman and
  Hussain}{2018}]{nuruzzaman2018survey}
M.~Nuruzzaman and O.~K. Hussain.
\newblock A survey on chatbot implementation in customer service industry
  through deep neural networks.
\newblock In {\em 2018 IEEE 15th International Conference on e-Business
  Engineering (ICEBE)}, pages 54--61. IEEE, 2018.

\bibitem[\protect\citeauthoryear{Okuda and Shoda}{2018}]{okuda2018ai}
Takuma Okuda and Sanae Shoda.
\newblock Ai-based chatbot service for financial industry.
\newblock {\em Fujitsu Scientific and Technical Journal}, 54(2):4--8, 2018.

\bibitem[\protect\citeauthoryear{Rosenfeld and
  Kraus}{2018}]{rosenfeld2018predicting}
Ariel Rosenfeld and Sarit Kraus.
\newblock Predicting human decision-making: From prediction to action.
\newblock {\em Synthesis Lectures on Artificial Intelligence and Machine
  Learning}, 12(1):1--150, 2018.

\bibitem[\protect\citeauthoryear{Rosenfeld \bgroup \em et al.\egroup
  }{2016}]{rosenfeld2016online}
A.~Rosenfeld, J.~Keshet, C.~V. Goldman, and S.~Kraus.
\newblock Online prediction of exponential decay time series with human-agent
  application.
\newblock In {\em ECAI 2016}, pages 595--603. IOS Press, 2016.

\bibitem[\protect\citeauthoryear{Rosenfeld \bgroup \em et al.\egroup
  }{2017}]{rosenfeld2017intelligent}
A.~Rosenfeld, N.~Agmon, O.~Maksimov, and S.~Kraus.
\newblock Intelligent agent supporting human--multi-robot team collaboration.
\newblock {\em AIJ}, 252:211--231, 2017.

\bibitem[\protect\citeauthoryear{Seker}{2014}]{seker2014using}
Alper Seker.
\newblock Using outputs of nasa-tlx for building a mental workload expert
  system.
\newblock {\em Gazi University Journal of Science}, 27(4):1131--1142, 2014.

\bibitem[\protect\citeauthoryear{Serban \bgroup \em et al.\egroup
  }{2015}]{serban2015}
I.~V. Serban, R.~Lowe, P.~Henderson, L.~Charlin, and J.~Pineau.
\newblock A survey of available corpora for building data-driven dialogue
  systems.
\newblock {\em arXiv preprint arXiv:1512.05742}, 2015.

\bibitem[\protect\citeauthoryear{Weizenbaum}{1966}]{eliza}
Joseph Weizenbaum.
\newblock Eliza—a computer program for the study of natural language
  communication between man and machine.
\newblock {\em Commun. ACM}, 9(1):36–45, January 1966.

\bibitem[\protect\citeauthoryear{Williams and Young}{2007}]{Williams2007}
J.D. Williams and S.~Young.
\newblock Partially observable markov decision processes for spoken dialog
  systems.
\newblock {\em COMPUT. SPEECH. LANG.}, 21(2):393--422, 2007.

\bibitem[\protect\citeauthoryear{Wong \bgroup \em et al.\egroup
  }{2018}]{Wong2018}
J.~Wong, L.~Hastings, K.~Negy, A.~J Gonzalez, S.~Onta{\~n}{\'o}n, and Y.~Lee.
\newblock Machine learning from observation to detect abnormal driving behavior
  in humans.
\newblock In {\em The 31st International Flairs Conference}, 2018.

\end{thebibliography}

\end{document}